\documentclass[letterpaper, 10 pt, conference]{ieeeconf}  

\IEEEoverridecommandlockouts                              

\usepackage{amsmath}
\usepackage{amssymb}
\usepackage[ruled, vlined]{algorithm2e}
\usepackage{booktabs}
\usepackage{cite}
\usepackage{graphicx}
\usepackage[colorlinks = true]{hyperref}
\usepackage{bookmark}
\usepackage{tablefootnote}
\usepackage[utf8]{inputenc}

\DeclareMathOperator*{\argmax}{arg\,max}

\title{\LARGE \bf
What My Motion tells me about Your Pose: \\
A Self-Supervised Monocular 3D Vehicle Detector 
}

\author{Cédric Picron$^{1}$, Punarjay Chakravarty$^2$, Tom Roussel$^1$, Tinne Tuytelaars$^1$%
\thanks{$^1$ ESAT-PSI, KU Leuven}
\thanks{$^2$ Ford Greenfield Labs, Palo Alto (this research was conducted as part of the Ford-KUL University Alliance program)}
}

\begin{document}
\maketitle
\thispagestyle{empty}
\pagestyle{empty}

\begin{abstract}
The estimation of the orientation of an observed vehicle relative to an Autonomous Vehicle (AV) from monocular camera data is an important building block in estimating its 6 DoF pose. Current Deep Learning based solutions for placing a 3D bounding box around this observed vehicle are data hungry and do not generalize well. In this paper, we demonstrate the use of monocular visual odometry for the self-supervised fine-tuning of a model for orientation estimation pre-trained on a reference domain. Specifically, while transitioning from a virtual dataset (vKITTI) to nuScenes, we recover up to $\mathbf{70}$\% of the performance of a fully supervised method. We subsequently demonstrate an optimization-based monocular 3D bounding box detector built on top of the self-supervised vehicle orientation estimator without the requirement of expensive labeled data. This allows 3D vehicle detection algorithms to be self-trained from large amounts of monocular camera data from existing commercial vehicle fleets.
\end{abstract}

\begin{figure*}
    \centering
    \includegraphics[scale=0.9]{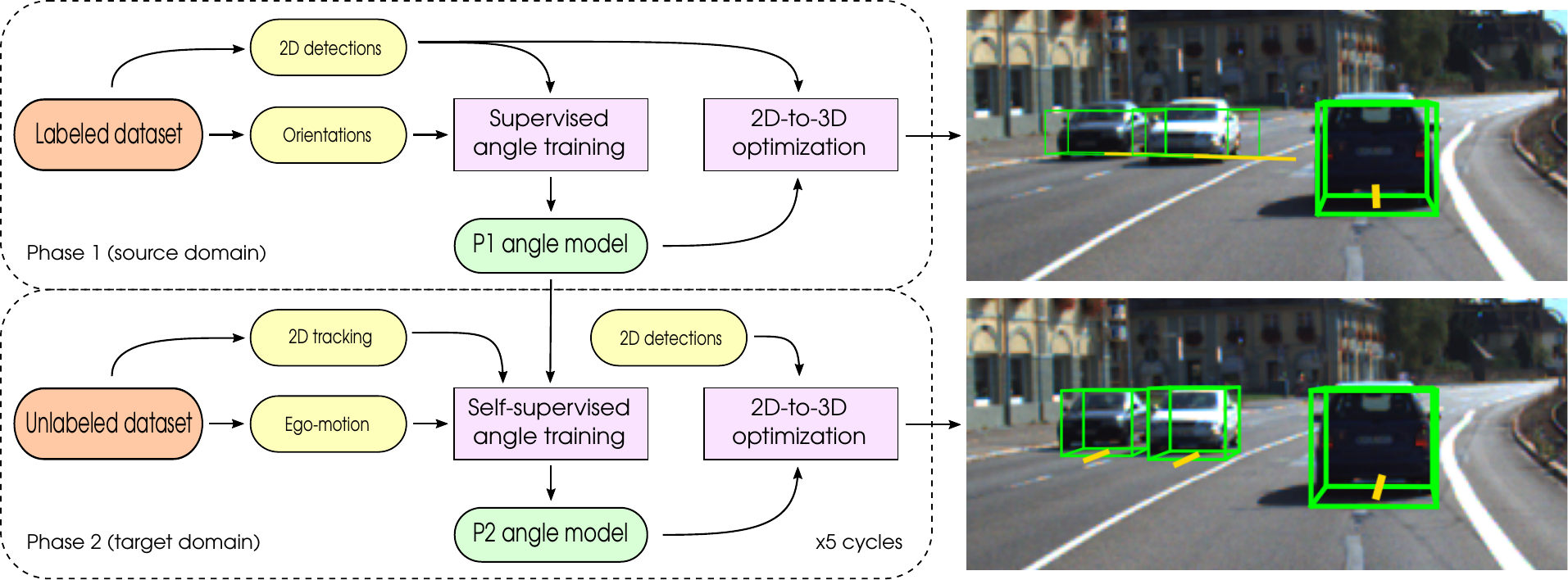}
    \caption{During Phase~1 (P1, \textit{top left}), the angle estimator is trained on the labeled source domain giving P1 model~$\mathcal{M}_0$. During cycle~$i$ of Phase~2 (P2, \textit{bottom left}), the P2 model $\mathcal{M}_i$ is obtained by fine-tuning $\mathcal{M}_{i-1}$ on the unlabeled target domain. We typically train for $5$ P2 cycles. A 2D-to-3D optimization process (\textit{middle}) is subsequently used to obtain 3D bounding boxes from 2D detections and yaw angle estimations. When evaluating the 3D bounding boxes on a target domain validation image (\textit{right}), we clearly see improved 3D boxes after self-supervised training of the angle estimator (\textit{bottom right}) compared to before (\textit{top right}).}
    \vspace*{-0.75cm} 
    \label{fig:overview}
\end{figure*}

\section{Introduction}
Despite substantial progress in orientation estimation and monocular 3D bounding box detection based on Deep Learning \cite{chen2016monocular, mousavian20173d, brazil2019m3d, ding2020learning}, major issues remain, including a strong dependence on expensive manual annotation and a lack of generalization to new settings. Combined, these
issues make regular fine-tuning or retraining of the models a necessity.

If one were able to use unlabeled data from monocular cameras to self-supervise this fine-tuning (\textit{i.e.}~without needing annotations), one would be able to leverage the large amounts of unlabeled data available from existing vehicle fleets on city streets to improve the task of orientation estimation and 3D bounding box detection in general. 

The estimation of ego-motion from a single camera is possible using monocular Visual Odometry (VO) or Visual SLAM; techniques that are relatively mature, given the progress of the last several years \cite{engel2014lsd, mur2015orb, mur2017orb}. These vSLAM/VO methods allow us to recover the path of the ego-vehicle, \textit{i.e.} the ego-motion, from a forward facing camera in global coordinates. When combined with detections of parked/stationary car instances, this information can be extremely valuable. As the stationary vehicle is fixed in global coordinates, orientation angle differences in the ego-frame between two time instances of the stationary car, can be inferred from ego-motion.

In this work, we leverage this ego-motion consistency for the self-supervised fine-tuning of our orientation estimator, which consists of a simple neural network operating on a cropped 2D detection of the vehicle. Our network is able to self-supervise its training, using monocular VO and image sequences of parked cars as the ego-vehicle drives past them. At inference time, our vehicle is able to take a single cropped image of a vehicle \textit{stationary or moving}, and output its orientation relative to the ego-vehicle.

A useful demonstration of our self-supervised method is in the Domain Adaptation (DA) \cite{wang2018deep} setting. DA comprises of techniques that help a neural network generalize to a different environment compared to the environment it was originally trained on. In our experiments, we start from a virtual dataset (vKITTI) \cite{gaidon2016virtual} and show consistent improvements in vehicle orientation estimation on monocular camera data from suburban Germany (KITTI) \cite{geiger2012we} and the cities of Boston and Singapore (nuScenes) \cite{caesar2020nuscenes}. Finally, we demonstrate an optimization based technique that uses a 2D object detector and the above self-supervised orientation angle estimator to estimate the full 3D bounding box of detected vehicles. An overview of our system is shown in Figure~\ref{fig:overview}. Thus, we show that we can go from simulated data to full monocular 3D bounding box detection, crucial to the perception stack of AVs, without the need for expensive annotations.

\section{Related Work}
\subsubsection{Monocular vehicle orientation estimation} With the popularization of deep learning in computer vision came the first monocular orientation estimators \cite{chen2016monocular, mousavian20173d, chabot2017deep}. Of particular interest to us, is the angle representation for the vehicle orientation. While early works \cite{chen2016monocular, chabot2017deep} relied on the global yaw angle as measured from the ego-frame, a decomposition in ray and local angle was found more beneficial \cite{mousavian20173d} (see also Figure~\ref{fig:sequence}). This was again emphasized in \cite{kundu20183d} where the difference between the allocentric and egocentric orientation was explained, which are merely synonyms for local and global yaw angles respectively. This decomposition has been used by multiple works ever since \cite{manhardt2019roi, liu2019deep, simonelli2019disentangling}. In some works, the orientation is found implicitly, by  localizing the 8~bounding box corners \cite{qin2019monogrnet}. Finally, some works represent the 3D vehicle orientation as a quaternion~\cite{simonelli2019disentangling}.

\subsubsection{Self-supervision in computer vision} 
Among the vast amount of unsupervised methods  \cite{kolesnikov2019revisiting}, we focus on self-supervised methods, \textit{i.e.} methods that make use of a so-called pretext task. Many such pretext tasks have been formulated, both for still images \cite{gidaris2018unsupervised, dosovitskiy2014discriminative, noroozi2016unsupervised, doersch2015unsupervised, wu2018unsupervised, kolesnikov2019revisiting} and videos \cite{goroshin2015unsupervised, misra2016shuffle, pathak2017learning, wang2015unsupervised}. Most similar to our work are \cite{agrawal2015learning, jayaraman2015learning, zhou2017unsupervised} that also use ego-motion, but instead use it as a representation learning tool \cite{agrawal2015learning, jayaraman2015learning} or for single image depth estimation \cite{zhou2017unsupervised}.

\subsubsection{Domain Adaptation in computer vision} Domain Adaptation (DA) is a longstanding research topic in computer vision \cite{csurka2017domain}.
\cite{donahue2013deep} were one of the first to observe that the DA problem persists for deeper networks. Since, many deep unsupervised DA methods have been devised. A simple DA method \cite{chang2019domain} consists of adapting the batch normalization statistics for each environment. For the autonomous driving setting, the DA methods have mostly focused on 2D semantic segmentation \cite{saleh2018effective, tsai2018learning} and 2D object detection \cite{chen2018domain, he2019multi, khodabandeh2019robust, rodriguez2019domain}. Many of these methods rely either on synthetic data or make use of adversarial feature learning \cite{ganin2016domain}. Recently, the DA task for full 3D object detection has also been considered in \cite{wang2020train} for one of the first times.

\subsubsection{3D bounding box optimization} We use a 3D bounding box optimization process to obtain a 3D box from a 2D detector and its yaw angle estimate, based on geometrical constraints. Similar constraints and optimization methods are found in \cite{mousavian20173d, hu2019joint, brazil2019m3d}.

\section{Method}
We introduce our method with definitions of local and global angles. This is followed by a detailed discussion about using monocular SLAM to self-supervise the training of the neural net that outputs the orientation angle of the observed vehicle. Next we describe the optimization approach to determine its full 3D bounding box.

\begin{figure}
    \centering
    \includegraphics[width=0.99\linewidth]{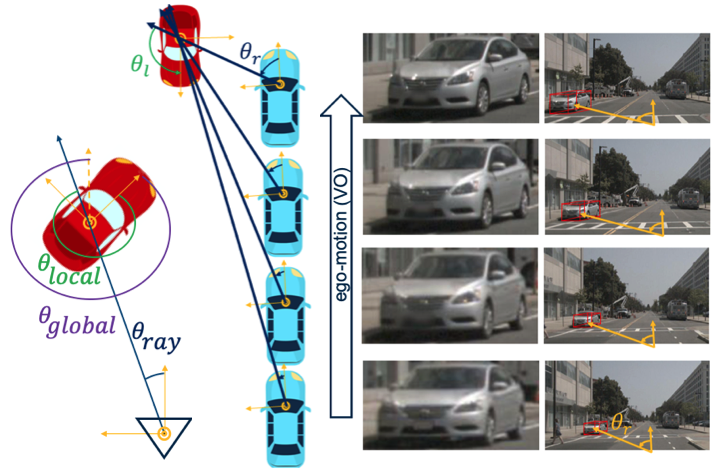}
    \caption{Left: Ray and local angles combine to make the global orientation (yaw) angle between ego and observed vehicle coordinate frames. Right: Ray angle increases and local angle decreases  as the ego-vehicle drives past the parked observed vehicle. This sequence of cropped and full images are used to compute local and ray angles respectively. Best viewed digitally.}
    \vspace*{-0.65cm} 
    \label{fig:sequence}
\end{figure}

\textbf{Global and local orientation angles:}
In determining the 6~DoF pose of the observed vehicle relative to the  ego-vehicle, an important parameter is the yaw angle. Roll and pitch angles remain relatively constant and close to zero for most street and highway conditions, assuming local planarity. This global relative yaw angle $\theta_g$, between the coordinate frames of the ego and observed vehicles was further decomposed to ray and local angles ($\theta_r$ and $\theta_l$) by \cite{mousavian20173d}. Here, the ray extending out from the camera coordinate frame intersecting with the centre of the observed vehicle coordinate frame (3D bounding box) subtends a ray angle determined by the optics of the observing camera (the camera intrinsics) and the centre of the 2D bounding box of the vehicle (left, Figure~\ref{fig:sequence}). A second, local angle, is then formed by the rotation of the observed vehicle about its own vertical axis, coinciding with the intersection of the ray and the observed vehicle coordinate frame. These ray and local angles add up to the global yaw angle by $\theta_g = \theta_r + \theta_l$.

Let us consider the ego-vehicle (blue) driving past another vehicle (red), as shown in Figure \ref{fig:sequence} (right). The appearance of the observed vehicle in the 2D image crop changes over time as one passes it, because of changes in $\theta_l$. Simultaneously, the position of the 2D bounding box in the image changes, leading to a change in $\theta_r$. $\theta_r$ can be obtained analytically from the 2D bounding box location and camera intrinsics. For $\theta_l$ we train a neural network starting from the appearance of the cropped vehicle inside the 2D bounding box. 

\textbf{Self-supervised training based on ego-motion:} We exploit ego-motion estimation (obtained from vSLAM) and the abundance of parked cars on public streets to self-supervise the training of the neural network that regresses the local angle of the vehicle from its cropped image. As we pass a stationary car, we obtain two snapshots of it, along with our ego-poses. The global angle difference between these ego-poses can be converted to a local angle difference for the observed car, which can then be used as self-supervisory signal to train the network. Concretely, consider a pair of global yaw angles $\theta_{slam}(t_i)$ and $\theta_{slam}(t_j)$ obtained from SLAM at time instances $t_i$ and $t_j$. \begin{equation}
    \Delta \hat{\theta}_g(t_i, t_j) = -\Delta \theta_{slam}(t_i, t_j)  = \theta_{slam}(t_i) - \theta_{slam}(t_j), \label{eq:reverse}
\end{equation}
where the minus sign is due to the fact that when the ego-vehicle turns left in the fixed SLAM reference frame, a stationary car appears to turn right as seen from the ego-frame. This global angle difference $\Delta \hat{\theta}_g(t_i, t_j)$ is then converted to a local angle difference $\Delta \hat{\theta}_l(t_i, t_j)$, by subtracting the respective ray angle difference $\Delta \theta_r(t_i, t_j)$ from it. Recall that the global angle is a sum of ray and local angles and that the ray angle can be determined analytically from the 2D bounding box and the camera intrinsics.

The hat symbol on the global angle difference is used to indicate the fact that the differences are estimates due to the approximate nature of the yaw angles (and the fact that the ray obtained from the centre of the 2D bounding box does not always intersect with the centre of the observed vehicle coordinate frame).

Note that the obtained global angles from SLAM describe the orientation of the ego-vehicle, independent of its translation or followed path. Scale ambiguities related to 3D paths of ego-vehicles obtained by monocular SLAM are therefore no problem here.

\textbf{Self-Supervised Training with a Siamese Net:} We experimented with feeding pairs of 2D vehicle crops into a Siamese net \cite{bromley1994signature}, which is expected to output local angle estimates for each of the cropped images. We trained this model stimulating equality between the predicted local angle difference $\Delta \tilde{\theta}_l(t_j, t_i)$ and the SLAM-based target difference $\Delta \hat{\theta}_l(t_i, t_j)$ based on the loss
\begin{equation}
    \mathcal{L} = d \left(\Delta \tilde{\theta}_l(t_j, t_i), \,\Delta \hat{\theta}_l(t_i, t_j) \right),
\end{equation}
with $d(x, y)$ a distance metric. However, this approach did not work in practice, and this is best explained with an example.  Suppose the ground-truth angles at two time instances are $10^\circ$ and $40^\circ$ while the network predicts $0^\circ$ and $20^\circ$ respectively. As in practice only the angle difference of $30^\circ$ is known, the network is told that the difference of $20^\circ$ between its predictions is too small and should be increased. Hence the network is told to mistakenly decrease its $0^\circ$ prediction and to correctly increase its $20^\circ$ prediction. On average, we observed that $40\%$ of the gradients flow in the wrong direction. This shows the difficulty of learning from differences: it is unknown which of the two predictions should be trusted more. To overcome this issue, we design a more robust approach described in the following subsections.

\textbf{The Domain Adaptation setting:} \label{sec:offset}
We consider the Domain Adaptation (DA) setting, where we assume we have a \textit{rough} angle estimator $\mathcal{M}_0$, perhaps trained using a small amount of ground-truth data or trained with data from another city or from virtual data. We desire the self-supervised fine-tuning of this network to a new dataset, to arrive at a Domain-Adapted network $\mathcal{M}_1$.

\textbf{Self-Supervised Target Angles:}
Consider $N$ consecutive detections of the same stationary car, from which following two sequences are derived: 1) a sequence of global angle measurements based on SLAM $s_n = -\theta_{slam}(t_n)$ and 2) a sequence of \textit{rough} global angle estimates $r_n$ based on $\mathcal{M}_0$. When estimating the stationary vehicle angle sequence, both sequences have flaws: 1) the $s_n$ sequence is only correct up to an unknown bias and 2) the $r_n$ contains various outliers despite having a meaningful bias. In what follows, we will therefore design a robust method to estimate the $s_n$ sequence bias~$b$ based on the $r_n$ sequence.



Consider the sequence of differences $d_n = r_n - s_n$. Possible approaches to estimate $b$ would be to compute its mean or median. Both approaches fail however as the former is sensitive to outliers and the latter assumes the outliers are symmetrically distributed around the median. We also tried off-the-shelf robust estimators 1) Huber regression \cite{huber2004robust} 2) RANSAC \cite{bolles1981ransac} and 3) the Theil-Sen regressor \cite{dang2008theil} as implemented in \cite{scikit-learn}, yet none of them were yielding satisfactory results. Instead, we therefore design our own robust estimator based on 1) outlier removal (named \textit{sequence pruning}) and 2) sequence removal.

For \textbf{sequence pruning}, we design an iterative process to prune outliers 1-by-1 from $d_n$ until a stop condition is satisfied. Given a set of $n$~indices $\mathcal{S}_n$, its most inconsistent index $i_{\textrm{max}}$ (\textit{i.e. outlier index}) is found by
\begin{equation}
    i_{\textrm{max}} = \argmax_{i \in \mathcal{S}_n} \mathcal{I}_i = \argmax_{i \in \mathcal{S}_n} \sum_{j \in \mathcal{S}_n} |d_j-d_i|. \label{eq:max}
\end{equation}
Next, its most consistent index $i_{\textrm{min}}$ is computed in a similar way. If now following condition is satisfied
\begin{equation}
    \frac{\mathcal{I}_{i_{\textrm{max}}}}{\mathcal{I}_{i_{\textrm{min}}}} > t_p \label{eq:condition}
\end{equation}
for a predetermined \textit{pruning threshold}~$t_p$, we prune entry $i_{\textrm{max}}$ from $\mathcal{S}_n$ and repeat the same process for the new set $ \mathcal{S}_{n-1} = \mathcal{S}_n \setminus \{i_{\textrm{max}}\}$. When starting from $\mathcal{S}_N = \{1, \dots, N\}$, this procedure is repeated until the condition \eqref{eq:condition} is no longer satisfied or when only two entries remain within the set. Finally, the estimated bias $\hat{b}$ is computed as the mean of $d_n$ with indices from the pruned set, yielding the self-supervised angle sequence $\theta_{\textrm{self}}(t_n) = s_n + \hat{b}$. 

We add a small post-processing step called \textbf{sequence removal}, where full sequences are removed from self-supervised training based on
\begin{equation}
        \sum_{i,j \in \mathcal{S}_3}|d_j - d_i| > 6 t_r,
\end{equation}
with $\mathcal{S}_3$ the set containing the three most consistent indices and with $t_r$ a predetermined \textit{remove threshold}. This absolute condition estimates that the inconsistencies between $s_n$ and $r_n$ (measured as $d_n$) are too large to estimate a reliable bias~$b$.


\textbf{Iterative Re-computation:} Once the self-supervised targets~$\theta_{\textrm{self}}(t_n)$ computed, we proceed by fine-tuning the initial model $\mathcal{M}_0$ with these targets yielding the model $\mathcal{M}_1$. Instead of stopping here, we can repeat the same process by computing improved self-supervised targets with $\mathcal{M}_1$ followed by fine-tuning with these, yielding the $\mathcal{M}_2$ model. In general, this iterative process called \textbf{cycling} can be repeated as many times as desired where in cycle~$i$ model~$\mathcal{M}_{i-1}$ is further fine-tuned to model~$\mathcal{M}_i$.

\textbf{Leaving Moving Vehicles in:} Above, we made the assumption that the observed vehicle is stationary, as only in this case the $s_n$ sequence is fully correct up to an unknown bias~$b$. In theory, moving cars should therefore be removed. In practice however, we observe that leaving them in does not hurt performance, obviating the need for a stationary/moving car detector. This can be understood by realizing that 1) sequence removal removes sequences inconsistent with ego-motion and 2) moving cars often move in straight lines yielding stationary global angles for which above $s_n$ sequence assumption is still valid.

\textbf{3D Bounding Box Optimization:} Starting from a 2D car detection and its yaw angle (estimated using the method described so far), we estimate its full 3D bounding box by optimizing the depth or distance to the vehicle using the difference between the 2D projection of the 3D bounding box and the 2D vehicle detection. Given a 2D detection, we place the center of a 3D box on the ray defined by the center of the 2D detection and the camera intrinsics. This 3D box is sized (length/width/height) using the median over vehicle statistics, oriented according the estimated yaw angle and placed at an initial arbitrary fixed distance away from the camera ($30$m). We then use the Nelder-Mead optimizer \cite{nelder1965simplex} to solve following optimization problem:
\begin{equation}
    Z^* = \argmax_{Z}\,\textrm{IoU}_{2D}(\mathcal{B}_{2D}, \,\mathcal{P}(\mathcal{B}_{3D}(Z))),
    \label{eq:optimization}
\end{equation}
with $\mathcal{B}_{2D}$ the estimated 2D box, $\mathcal{P(\cdot)}$ the 3D to 2D projection operator and $\mathcal{B}_{3D}(Z)$ the estimated 3D box which varies according to the given depth~$Z$. The resulting depth~$Z^*$ is then used to yield the 3D bounding box of the corresponding 2D car detection. Note that the box size (width, height, length) is not being optimized as it renders the optimization unstable due to its tendency to reduce the 3D box to a 2D box (\textit{i.e.}~length or width get collapsed to zero), which is not penalized by the 2D IoU metric. This could be avoided by a segmentation based silhouette metric~\cite{wang2020directshape} and is left as future work.


\section{Experiments} \label{sec:experiments}
\textbf{Datasets:} We use images from a single forward facing camera from 3 self-driving datasets:  KITTI~\cite{geiger2012we}, nuScenes~\cite{caesar2020nuscenes} and Virtual KITTI~\cite{gaidon2016virtual} (vKITTI), where vKITTI is a virtual clone of the famous KITTI dataset. Our Phase 1 (P1) model $\mathcal{M}_0$ for orientation angle estimation is obtained by fully supervised training on the source dataset. The Phase 2 (P2) angle estimation models $\mathcal{M}_i$ are obtained after self-supervised fine-tuning of $\mathcal{M}_{i-1}$ on the target dataset. For KITTI, we use the \textit{tracking} version of the dataset for orientation angle experiments, and the \textit{object detection} version for the 3D bounding box experiments\footnote{Note that some images in the KITTI validation set might coincide with their virtual counterparts from vKITTI used for training, leading to overoptimistic results for the vKITTI to KITTI domain transfer (VK $\rightarrow$ K).}.

\textbf{Evaluation metrics:} For the orientation angle experiments, we split the data into an 80/20 train/test split, during both the P1 and P2 training phases. We use the median orientation angle error to evaluate its accuracy. 

For the 3D bounding box experiments, we use all available training/test data of the target domain. We use both the median 3D component errors (\textit{i.e.} translation, orientation and box size errors) and the KITTI AP's as evaluation metrics.

\textbf{Baselines:} For the orientation angle experiments, we experimented with two Domain Adaptation (DA) algorithms~\cite{ganin2016domain, chang2019domain} for adapting P1 to P2.  Note that the different camera intrinsics are already compensated for in the transformation to local angles, so the domain differences are mostly visual. However, both of them gave worse or similar results for P2 compared to P1, on the target dataset. This is possibly because these methods were designed for classification tasks. We therefore use the performance of P1 on the target dataset (before fine-tuning) as our baseline.

No baselines are provided for the 3D object detection experiments, as no competitors exist on the self-supervised monocular 3D object detection task. Note that during these 3D object detection experiments, we only make use of angle estimators pre-trained on virtual data. Our method can therefore be considered as fully self-supervised without the use human annotations, which is in contrast to DA-based methods as in~\cite{wang2020train}.

\textbf{Implementation Details:} For our orientation angle estimation neural network, we use the ResNeXt-32x4d model \cite{xie2017aggregated} pre-trained on ImageNet and with its final fully connected layer replaced with a new linear layer. The network is trained to take in a cropped view of the vehicle, and outputs its orientation angle. During P1, this model is trained using ground-truth 2D detections and vehicle orientation values. During P2, 2D vehicle detections are obtained from a Mask R-CNN model with a ResNet-50 FPN backbone \cite{he2017mask} pre-trained on COCO, while the ego-motion of the vehicle is estimated with monocular ORB-SLAM2 using its default KITTI settings \cite{mur2017orb}. 2D detections between frames over time as the ego-vehicle passes the observed vehicle, are linked based on the ground-truth data associations in these experiments, but could be replaced by an off-the-shelf tracker \cite{weng20203d}. The pruning and removal threshold are set to $t_p = 1.0$ and $t_r = 1.0^\circ$ respectively across all experiments. We use the smooth-$L_1$ loss with a quadratic region of width~$20^\circ$ for training the angle estimator. The model is optimized using SGD with momentum~$0.9$ and weight decay~$10^{-4}$. We use a batch size of $32$ with a learning rate of $2\cdot10^{-5}$ (where losses within a batch are summed instead of averaged). P1 is trained for 30 epochs, with learning rate decrease with a factor $10$ after $20$ epochs. P2 is trained for $5$ cycles of iterative self-supervision using both stationary and moving cars. Each of these cycles is trained for $30$ epochs on the KITTI dataset, with a learning rate decrease after $20$ epochs, and for $10$ epochs on the nuScenes dataset, with a learning rate decrease after $7$ epochs. Random horizontal flipping of the cropped image is used as data augmentation during both training phases P1 and P2.

\subsection{Orientation angle estimation}
\textbf{Main results:} Table~\ref{tab:main} shows the results (in terms of median angle error) of our self-supervised angle estimator bridging the domain gap across datasets, compared to two baselines that serve as a lower and upper bound respectively and comprise the first two rows of the table: a pre-trained model (P1) on the target dataset (on which we hope to improve upon) and a fully supervised model P2 on the target dataset (whose performance we hope to edge closer to). In the subsequent 5 rows, we show the results of our iterative orientation angle training, with each row (cycle) being trained using the P2 model from the previous row for self-supervision. The columns indicate the datasets we are trying to bridge the gap between; VK$\rightarrow$N means that a model P1 pre-trained on vKITTI is being fine-tuned for nuScenes. For VK$\rightarrow$N, we start from a P1 error of $41.2^\circ$ and decrease this right down to $13.9^\circ$ through 5 cycles of self-supervised fine-tuning. This is a gain of approximately {\boldmath$70\%$} (compared to supervised) starting from a virtual dataset and requiring no manual annotations whatsoever. We see a similar benefit from our iterative self-supervised training for K$\rightarrow$N. However, for the two domain shifts to the right of the figure: VK$\rightarrow$K and N$\rightarrow$K, the results are less promising. For the first of these two cases, we manage to decrease the error from $14.9^\circ$ to $10.4^\circ$ using the first cycle of self-supervision, but subsequent cycles yield no additional gain. For the second, our P2 error remains about the same as the P1 error of $4.2^\circ$, with no gain achieved from self-supervised domain adaptation. We conclude that our method of self-supervised fine-tuning is effective only if the P1 model corresponding to a particular domain change has an error larger than ${\sim} 10^\circ$ to begin with. Our supplementary video submission shows qualitative results and demonstrates how our final P2 angle estimators benefited from our self-supervised fine-tuning scheme on both KITTI and nuScenes.

\begin{table}[t]
    \caption[caption]{Median orientation angle errors \\ (VK = Virtual KITTI, K = KITTI and N = nuScenes)}
    \centering
    \setlength{\tabcolsep}{8pt}
    \begin{tabular}{c c c|c c}
        \toprule
        & VK $\rightarrow$ N & K $\rightarrow$ N & VK $\rightarrow$ K & N $\rightarrow$ K\\
        \midrule
        P1 model & $41.2^\circ$ & $25.1^\circ$ & $14.9^\circ$ & $4.2^\circ$ \\
        P2 (supervised) & $2.7^\circ$ & $2.7^\circ$ & $2.9^\circ$ & $2.9^\circ$ \\
        \midrule
        P2 (cycle 1) & $23.1^\circ$ & $14.1^\circ$ & $10.4^\circ$ & $4.4^\circ$ \\
        P2 (cycle 2) & $17.2^\circ$ & $12.0^\circ$ & $9.7^\circ$ & $4.4^\circ$ \\
        P2 (cycle 3) & $15.2^\circ$ & $12.0^\circ$ & $10.2^\circ$ & $4.3^\circ$ \\
        P2 (cycle 4) & $14.2^\circ$ & $11.3^\circ$ & $10.2^\circ$ & $4.3^\circ$ \\
        P2 (cycle 5) & $13.9^\circ$ & $10.9^\circ$ & $10.7^\circ$ & $4.2^\circ$ \\
        \bottomrule
    \end{tabular}
    \label{tab:main}
\end{table}

\begin{table}[t]
    \caption{Orientation angle ablation study}
    \centering
    \setlength{\tabcolsep}{7.5pt}
    \begin{tabular}{c c c|c c}
        \toprule
        & VK $\rightarrow$ N & K $\rightarrow$ N & VK $\rightarrow$ K & N $\rightarrow$ K \\
        \midrule
        P1 model & $41.2^\circ$ & $25.1^\circ$ & $14.9^\circ$ & $4.2^\circ$ \\
        P2 (supervised) & $2.7^\circ$ & $2.7^\circ$ & $2.9^\circ$ & $2.9^\circ$ \\
        P2 (cycle 1) & $23.1^\circ$ & $14.1^\circ$ & $10.4^\circ$ & $4.4^\circ$ \\
        \midrule
        No pruning & $37.9^\circ$ & $25.0^\circ$ & $28.9^\circ$ & $10.0^\circ$ \\
        No removal & $31.0^\circ$ & $16.9^\circ$ & $13.0^\circ$ & $4.6^\circ$ \\
        \midrule
        Gt. detections & $23.5^\circ$ & $13.6^\circ$ & $10.6^\circ$ & $4.1^\circ$ \\
        Gt. ego-motion\tablefootnote{The ablation uses both ground-truth ego-motion and also accounts for moving vehicles orientation changes. The influence of the latter is minimal as suggested by the \textit{no moving cars} ablation.} & $24.0^\circ$ & $14.8^\circ$ & $8.6^\circ$ & $3.5^\circ$ \\
        No moving cars & $23.8^\circ$ & $14.4^\circ$ & - & - \\
        \bottomrule
    \end{tabular}
    \vspace*{-0.20cm} 
    \label{tab:ablation}
\end{table}

\begin{table}[t]
    \caption[caption]{Median errors for different 3D components \\ (K = KITTI and N = nuScenes)}
    \centering
    \resizebox{\columnwidth}{!}{
    \begin{tabular}{l c c|c c}
        \toprule
        & K (P1) & K (P2) & N (P1) & N (P2)\\
        \midrule
        Height error (m) & \multicolumn{2}{c|}{$0.08$}  & \multicolumn{2}{c}{$0.13$}  \\
        Length error (m) & \multicolumn{2}{c|}{$0.30$}  & \multicolumn{2}{c}{$0.72$}  \\
        Width error (m) & \multicolumn{2}{c|}{$0.06$}  & \multicolumn{2}{c}{$0.41$}  \\
        \midrule
        X error (m) & $0.78$ & $0.62$ & $0.91$ & $0.70$ \\
        Y error (m) & $0.21$ & $0.17$ & $0.19$ & $0.17$ \\
        Z error (m) & $3.46$ & $2.89$ & $4.72$ & $3.73$ \\
        \midrule
        Yaw error ($^\circ$) & $19.8$ & $10.6$ & $31.0$ & $7.2$ \\
        \bottomrule
    \end{tabular}}
    \vspace*{-0.60cm} 
    \label{tab:errors}
\end{table}

\begin{table*}[t]
    \caption{AP's as defined by the KITTI benchmark }
    \centering
    \setlength{\tabcolsep}{6pt}
    \begin{tabular}{l|c c c|c c c|c c c}
        \toprule
        & & \textbf{2D detection} & & & \textbf{Bird's eye view} & & & \textbf{3D detection} & \\ 
        & Easy & Moderate & Hard & Easy & Moderate & Hard & Easy & Moderate & Hard \\
        \midrule
        KITTI (P1) & $71.5$ & $66.3$ & $54.6$ & $0.39$ & $0.31$ & $0.23$ & $0.10$ & $0.07$ & $0.07$ \\
        KITTI (P2) & $71.5$ & $66.3$ & $54.6$ & $0.38$ & $0.37$ & $0.27$ & $0.10$ & $0.11$ & $0.07$ \\ 
        \midrule
        nuScenes (P1) & $18.8$ & $17.1$ & $17.1$ & $0.05$ & $0.05$ & $0.05$ & $0.01$ & $0.01$ & $0.01$ \\
        nuScenes (P2) & $18.8$ & $17.1$ & $17.1$ & $0.09$ & $0.09$ & $0.09$ & $0.03$ & $0.02$ & $0.02$ \\
        \bottomrule
    \end{tabular}
    \vspace*{-0.60cm} 
    \label{tab:AP}
\end{table*}

\textbf{Ablation Study:}  Next, we perform an ablation study of the different components and mechanisms of our method. Here we keep the same settings as before, except that we fine-tune for one cycle only. The results of the ablation study are found in Table~\ref{tab:ablation}. The first three rows are copied from Table~\ref{tab:main} which operate as reference against which our ablations will be compared. Rows $4$ and $5$ show the results of ablating the pruning and removal mechanisms: the orientation angle median errors increase compared to our fully equipped P2 model, unambiguously showing the importance of both components. Next, the effect of using our Mask-RCNN based 2D object detector is determined by comparing it to a ground-truth detector, and the results show no major performance hits in using it. The next row investigates the effect of using ego-motion estimation from monocular SLAM for estimating the angle differences, versus using ground-truth angle differences (obtained by using GPS + LIDAR SLAM). For adaptation to nuScenes from vKITTI and KITTI, we see no difference between using SLAM and ground-truth, but for adaptation to KITTI from vKITTI and nuScenes, we see a minor improvement. Finally, we look at removing moving cars from the dataset using the ground-truth, versus including them in the self-supervised training, and this again does not affect the performance while adapting to nuScenes. No moving/stationary labels are provided for KITTI and these entries are left blank.

\subsection{3D bounding box estimation}
We now report experiments on the full 3D bounding box estimation using our optimizer. We use P2 angle estimators (after 5 cycles) that were P1 pre-trained on vKITTI. Thus, we provide true self-supervision, where we start from a virtual dataset and show full 3D vehicle detection on the KITTI and nuScenes datasets. Table~\ref{tab:errors} compares the median 3D component errors of the estimated 3D bounding boxes using an angle estimator without (P1) and with (P2) fine-tuning. Error stats are calculated over 3D boxes with 2D detections matched with a ground-truth box with at least $0.5$ 2D IoU overlap. In general, we see decent improvements when using the fine-tuned P2 angle estimator compared to the P1 angle estimator, except for the box sizes which are the same in both cases. When looking more closely at the median errors with fine-tuned P2 angle estimator, we observe the following. The first 3 rows show that the median box size errors are relatively small - a median of about $30$ and $6$ cm along the length and width of the car for vehicles in KITTI and $72$ and $41$ cm in nuScenes respectively, despite using general vehicle size statistics. The next 3 rows show reasonable median distance errors of sub $1$m in the X and Y directions. The Z error (along the viewing direction of the forward facing camera and stretching out from the ego-vehicle into the direction of travel) is $2.9$m for KITTI and $3.7$m for nuScenes. This is less than the length of a mid-sized vehicle which is not bad, considering that some vehicles are more than $100$m away. The last row describes the yaw angle error (which as described earlier) is the sum of the ray angle (determined by the 2D bounding box and camera intrinsics) and the local angle estimated by our self-supervised neural net. Small differences are observed with the results from Table~\ref{tab:main} due to differences in evaluation datasets.

Next, we look at the Average Precision (AP) metrics as used in the KITTI benchmark \cite{geiger2012we}. Here, positive matches are considered when the 3D IoU between the predicted and estimated bounding boxes is at least $0.7$. In Table~\ref{tab:AP}, we report AP values for easy, moderate and hard detections (as defined by KITTI) for the 2D image plane detector, the 3D bounding box detections from our full algorithm and its Bird's Eye View (BEV) version by collapsing the 3D bounding box onto the X-Z plane. Note that for nuScenes, the moderate and hard categories contain the same detections as no occluded or truncated labels are provided.

The first 3 columns of Table~\ref{tab:AP} show the 2D object detection AP results on KITTI and nuScenes using a Mask R-CNN model pre-trained on COCO. This is an off-the-shelf 2D detector that we do not further fine-tune on the KITTI and nuScenes domains as this would be contrary to our self-supervised problem setting which prohibits the use of ground-truth data. These results are for illustrative purposes only. The remaining 6 columns show the AP results for both the BEV and the full 3D object detection tasks. Our 3D object detection AP metrics improve with self-supervised fine-tuning, but are still quite low compared to numbers like $19.7$ and $14.1$ for BEV and 3D Bounding Boxes on KITTI (moderate) using a SoTA 3D object detector \cite{brazil2020kinematic}. However, it is important to point out that our method is completely self-supervised, and theirs is a supervised method. On the other hand, our errors using the bounding box dimension and distance metrics (Table~\ref{tab:errors}) are quite reasonable (within single car length up to a detection distance of 100m) and it can be argued that these results are good enough for an AV to maintain a safe distance from in-path vehicles. Qualitatively too, we obtain pleasing 3D bounding box results as shown in the supplementary video submission. The method on average runs at $1.15$ FPS on a GTX 1050 with $28\%$, $55\%$ and $17\%$ of time spent on 2D detection, angle estimation and 3D bounding box optimization respectively. This work has not focused on optimizing the run-times of any part of the algorithm and this is left as future work.

\section{Conclusion}
This paper uses the ego-vehicle’s motion and the 2D tracking of cars to self-supervise the fine-tuning of a vehicle orientation estimator. The neural net outputs the observed vehicle's yaw angle with respect to the ego-vehicle, which is an important component in the estimation of its full 6 DoF pose. The potential of this approach is demonstrated in a Domain Adaptation (DA) setting, where supervised training (on a virtual dataset) is followed by self-supervised fine-tuning (on a real dataset), recovering up to $70\%$ of performance relative to fully supervised training. By leveraging simulated data from Virtual KITTI, we were able to obtain a median validation error of $13.9^\circ$ on nuScenes without using a single human annotation, significantly reducing the gap with supervised methods. Finally, we build a self-supervised monocular 3D object detector on top of our self-supervised angle estimator, yielding pleasing qualitative results while clearly improving upon the 3D object detector not using our self-supervised angle estimator. The global angle self-supervision obtained from the ego-vehicle motion is independent of scale. This allows a 3D object detector to be potentially fine-tuned using footage from the vast amounts of monocular camera data already available from existing commercial vehicle fleets. 

\bibliographystyle{IEEEtran}
\bibliography{references}

\begin{thebibliography}{10}
\providecommand{\url}[1]{#1}
\csname url@rmstyle\endcsname
\providecommand{\newblock}{\relax}
\providecommand{\bibinfo}[2]{#2}
\providecommand\BIBentrySTDinterwordspacing{\spaceskip=0pt\relax}
\providecommand\BIBentryALTinterwordstretchfactor{4}
\providecommand\BIBentryALTinterwordspacing{\spaceskip=\fontdimen2\font plus
\BIBentryALTinterwordstretchfactor\fontdimen3\font minus
  \fontdimen4\font\relax}
\providecommand\BIBforeignlanguage[2]{{%
\expandafter\ifx\csname l@#1\endcsname\relax
\typeout{** WARNING: IEEEtran.bst: No hyphenation pattern has been}%
\typeout{** loaded for the language `#1'. Using the pattern for}%
\typeout{** the default language instead.}%
\else
\language=\csname l@#1\endcsname
\fi
#2}}

\bibitem{chen2016monocular}
X.~Chen, K.~Kundu, Z.~Zhang, H.~Ma, S.~Fidler, and R.~Urtasun, ``Monocular 3d
  object detection for autonomous driving,'' in \emph{Proceedings of the IEEE
  Conference on Computer Vision and Pattern Recognition}, 2016, pp. 2147--2156.

\bibitem{mousavian20173d}
A.~Mousavian, D.~Anguelov, J.~Flynn, and J.~Kosecka, ``3d bounding box
  estimation using deep learning and geometry,'' in \emph{Proceedings of the
  IEEE Conference on Computer Vision and Pattern Recognition}, 2017, pp.
  7074--7082.

\bibitem{brazil2019m3d}
G.~Brazil and X.~Liu, ``M3d-rpn: Monocular 3d region proposal network for
  object detection,'' in \emph{Proceedings of the IEEE International Conference
  on Computer Vision}, 2019, pp. 9287--9296.

\bibitem{ding2020learning}
M.~Ding, Y.~Huo, H.~Yi, Z.~Wang, J.~Shi, Z.~Lu, and P.~Luo, ``Learning
  depth-guided convolutions for monocular 3d object detection,'' in
  \emph{Proceedings of the IEEE/CVF Conference on Computer Vision and Pattern
  Recognition Workshops}, 2020, pp. 1000--1001.

\bibitem{engel2014lsd}
J.~Engel, T.~Sch{\"o}ps, and D.~Cremers, ``Lsd-slam: Large-scale direct
  monocular slam,'' in \emph{European conference on computer vision}.\hskip 1em
  plus 0.5em minus 0.4em\relax Springer, 2014, pp. 834--849.

\bibitem{mur2015orb}
R.~Mur-Artal, J.~M.~M. Montiel, and J.~D. Tardos, ``Orb-slam: a versatile and
  accurate monocular slam system,'' \emph{IEEE transactions on robotics},
  vol.~31, no.~5, pp. 1147--1163, 2015.

\bibitem{mur2017orb}
R.~Mur-Artal and J.~D. Tard{\'o}s, ``Orb-slam2: An open-source slam system for
  monocular, stereo, and rgb-d cameras,'' \emph{IEEE Transactions on Robotics},
  vol.~33, no.~5, pp. 1255--1262, 2017.

\bibitem{wang2018deep}
M.~Wang and W.~Deng, ``Deep visual domain adaptation: A survey,''
  \emph{Neurocomputing}, vol. 312, pp. 135--153, 2018.

\bibitem{gaidon2016virtual}
A.~Gaidon, Q.~Wang, Y.~Cabon, and E.~Vig, ``Virtual worlds as proxy for
  multi-object tracking analysis,'' in \emph{Proceedings of the IEEE conference
  on computer vision and pattern recognition}, 2016, pp. 4340--4349.

\bibitem{geiger2012we}
A.~Geiger, P.~Lenz, and R.~Urtasun, ``Are we ready for autonomous driving? the
  kitti vision benchmark suite,'' in \emph{2012 IEEE Conference on Computer
  Vision and Pattern Recognition}.\hskip 1em plus 0.5em minus 0.4em\relax IEEE,
  2012, pp. 3354--3361.

\bibitem{caesar2020nuscenes}
H.~Caesar, V.~Bankiti, A.~H. Lang, S.~Vora, V.~E. Liong, Q.~Xu, A.~Krishnan,
  Y.~Pan, G.~Baldan, and O.~Beijbom, ``nuscenes: A multimodal dataset for
  autonomous driving,'' in \emph{Proceedings of the IEEE/CVF Conference on
  Computer Vision and Pattern Recognition}, 2020, pp. 11\,621--11\,631.

\bibitem{chabot2017deep}
F.~Chabot, M.~Chaouch, J.~Rabarisoa, C.~Teuliere, and T.~Chateau, ``Deep manta:
  A coarse-to-fine many-task network for joint 2d and 3d vehicle analysis from
  monocular image,'' in \emph{Proceedings of the IEEE conference on computer
  vision and pattern recognition}, 2017, pp. 2040--2049.

\bibitem{kundu20183d}
A.~Kundu, Y.~Li, and J.~M. Rehg, ``3d-rcnn: Instance-level 3d object
  reconstruction via render-and-compare,'' in \emph{Proceedings of the IEEE
  conference on computer vision and pattern recognition}, 2018, pp. 3559--3568.

\bibitem{manhardt2019roi}
F.~Manhardt, W.~Kehl, and A.~Gaidon, ``Roi-10d: Monocular lifting of 2d
  detection to 6d pose and metric shape,'' in \emph{Proceedings of the IEEE
  Conference on Computer Vision and Pattern Recognition}, 2019, pp. 2069--2078.

\bibitem{liu2019deep}
L.~Liu, J.~Lu, C.~Xu, Q.~Tian, and J.~Zhou, ``Deep fitting degree scoring
  network for monocular 3d object detection,'' in \emph{Proceedings of the IEEE
  Conference on Computer Vision and Pattern Recognition}, 2019, pp. 1057--1066.

\bibitem{simonelli2019disentangling}
A.~Simonelli, S.~R. Bulo, L.~Porzi, M.~L{\'o}pez-Antequera, and
  P.~Kontschieder, ``Disentangling monocular 3d object detection,'' in
  \emph{Proceedings of the IEEE International Conference on Computer Vision},
  2019, pp. 1991--1999.

\bibitem{qin2019monogrnet}
Z.~Qin, J.~Wang, and Y.~Lu, ``Monogrnet: A geometric reasoning network for
  monocular 3d object localization,'' in \emph{Proceedings of the AAAI
  Conference on Artificial Intelligence}, vol.~33, 2019, pp. 8851--8858.

\bibitem{kolesnikov2019revisiting}
A.~Kolesnikov, X.~Zhai, and L.~Beyer, ``Revisiting self-supervised visual
  representation learning,'' in \emph{Proceedings of the IEEE conference on
  Computer Vision and Pattern Recognition}, 2019, pp. 1920--1929.

\bibitem{gidaris2018unsupervised}
S.~Gidaris, P.~Singh, and N.~Komodakis, ``Unsupervised representation learning
  by predicting image rotations,'' \emph{arXiv preprint arXiv:1803.07728},
  2018.

\bibitem{dosovitskiy2014discriminative}
A.~Dosovitskiy, J.~T. Springenberg, M.~Riedmiller, and T.~Brox,
  ``Discriminative unsupervised feature learning with convolutional neural
  networks,'' in \emph{Advances in neural information processing systems},
  2014, pp. 766--774.

\bibitem{noroozi2016unsupervised}
M.~Noroozi and P.~Favaro, ``Unsupervised learning of visual representations by
  solving jigsaw puzzles,'' in \emph{European Conference on Computer
  Vision}.\hskip 1em plus 0.5em minus 0.4em\relax Springer, 2016, pp. 69--84.

\bibitem{doersch2015unsupervised}
C.~Doersch, A.~Gupta, and A.~A. Efros, ``Unsupervised visual representation
  learning by context prediction,'' in \emph{Proceedings of the IEEE
  International Conference on Computer Vision}, 2015, pp. 1422--1430.

\bibitem{wu2018unsupervised}
Z.~Wu, Y.~Xiong, S.~X. Yu, and D.~Lin, ``Unsupervised feature learning via
  non-parametric instance discrimination,'' in \emph{Proceedings of the IEEE
  Conference on Computer Vision and Pattern Recognition}, 2018, pp. 3733--3742.

\bibitem{goroshin2015unsupervised}
R.~Goroshin, J.~Bruna, J.~Tompson, D.~Eigen, and Y.~LeCun, ``Unsupervised
  learning of spatiotemporally coherent metrics,'' in \emph{Proceedings of the
  IEEE international conference on computer vision}, 2015, pp. 4086--4093.

\bibitem{misra2016shuffle}
I.~Misra, C.~L. Zitnick, and M.~Hebert, ``Shuffle and learn: unsupervised
  learning using temporal order verification,'' in \emph{European Conference on
  Computer Vision}.\hskip 1em plus 0.5em minus 0.4em\relax Springer, 2016, pp.
  527--544.

\bibitem{pathak2017learning}
D.~Pathak, R.~Girshick, P.~Doll{\'a}r, T.~Darrell, and B.~Hariharan, ``Learning
  features by watching objects move,'' in \emph{Proceedings of the IEEE
  Conference on Computer Vision and Pattern Recognition}, 2017, pp. 2701--2710.

\bibitem{wang2015unsupervised}
X.~Wang and A.~Gupta, ``Unsupervised learning of visual representations using
  videos,'' in \emph{Proceedings of the IEEE International Conference on
  Computer Vision}, 2015, pp. 2794--2802.

\bibitem{agrawal2015learning}
P.~Agrawal, J.~Carreira, and J.~Malik, ``Learning to see by moving,'' in
  \emph{Proceedings of the IEEE international conference on computer vision},
  2015, pp. 37--45.

\bibitem{jayaraman2015learning}
D.~Jayaraman and K.~Grauman, ``Learning image representations tied to
  ego-motion,'' in \emph{Proceedings of the IEEE International Conference on
  Computer Vision}, 2015, pp. 1413--1421.

\bibitem{zhou2017unsupervised}
T.~Zhou, M.~Brown, N.~Snavely, and D.~G. Lowe, ``Unsupervised learning of depth
  and ego-motion from video,'' in \emph{Proceedings of the IEEE Conference on
  Computer Vision and Pattern Recognition}, 2017, pp. 1851--1858.

\bibitem{csurka2017domain}
G.~Csurka, \emph{Domain adaptation in computer vision applications}.\hskip 1em
  plus 0.5em minus 0.4em\relax Springer, 2017, vol.~2.

\bibitem{donahue2013deep}
J.~Donahue, Y.~Jia, O.~Vinyals, J.~Hoffman, N.~Zhang, E.~Tzeng, and T.~Darrell,
  ``A deep convolutional activation feature for generic visual recognition,''
  \emph{arXiv preprint arXiv:1310.1531}, vol.~1, 2013.

\bibitem{chang2019domain}
W.-G. Chang, T.~You, S.~Seo, S.~Kwak, and B.~Han, ``Domain-specific batch
  normalization for unsupervised domain adaptation,'' in \emph{Proceedings of
  the IEEE Conference on Computer Vision and Pattern Recognition}, 2019, pp.
  7354--7362.

\bibitem{saleh2018effective}
F.~S. Saleh, M.~S. Aliakbarian, M.~Salzmann, L.~Petersson, and J.~M. Alvarez,
  ``Effective use of synthetic data for urban scene semantic segmentation,'' in
  \emph{European Conference on Computer Vision}.\hskip 1em plus 0.5em minus
  0.4em\relax Springer, 2018, pp. 86--103.

\bibitem{tsai2018learning}
Y.-H. Tsai, W.-C. Hung, S.~Schulter, K.~Sohn, M.-H. Yang, and M.~Chandraker,
  ``Learning to adapt structured output space for semantic segmentation,'' in
  \emph{Proceedings of the IEEE Conference on Computer Vision and Pattern
  Recognition}, 2018, pp. 7472--7481.

\bibitem{chen2018domain}
Y.~Chen, W.~Li, C.~Sakaridis, D.~Dai, and L.~Van~Gool, ``Domain adaptive faster
  r-cnn for object detection in the wild,'' in \emph{Proceedings of the IEEE
  conference on computer vision and pattern recognition}, 2018, pp. 3339--3348.

\bibitem{he2019multi}
Z.~He and L.~Zhang, ``Multi-adversarial faster-rcnn for unrestricted object
  detection,'' in \emph{Proceedings of the IEEE International Conference on
  Computer Vision}, 2019, pp. 6668--6677.

\bibitem{khodabandeh2019robust}
M.~Khodabandeh, A.~Vahdat, M.~Ranjbar, and W.~G. Macready, ``A robust learning
  approach to domain adaptive object detection,'' in \emph{Proceedings of the
  IEEE International Conference on Computer Vision}, 2019, pp. 480--490.

\bibitem{rodriguez2019domain}
A.~L. Rodriguez and K.~Mikolajczyk, ``Domain adaptation for object detection
  via style consistency,'' \emph{arXiv preprint arXiv:1911.10033}, 2019.

\bibitem{ganin2016domain}
Y.~Ganin, E.~Ustinova, H.~Ajakan, P.~Germain, H.~Larochelle, F.~Laviolette,
  M.~Marchand, and V.~Lempitsky, ``Domain-adversarial training of neural
  networks,'' \emph{The Journal of Machine Learning Research}, vol.~17, no.~1,
  pp. 2096--2030, 2016.

\bibitem{wang2020train}
Y.~Wang, X.~Chen, Y.~You, L.~E. Li, B.~Hariharan, M.~Campbell, K.~Q.
  Weinberger, and W.-L. Chao, ``Train in germany, test in the usa: Making 3d
  object detectors generalize,'' in \emph{Proceedings of the IEEE/CVF
  Conference on Computer Vision and Pattern Recognition}, 2020, pp.
  11\,713--11\,723.

\bibitem{hu2019joint}
H.-N. Hu, Q.-Z. Cai, D.~Wang, J.~Lin, M.~Sun, P.~Kr{\"a}henb{\"u}hl,
  T.~Darrell, and F.~Yu, ``Joint monocular 3d vehicle detection and tracking,''
  in \emph{ICCV}, 2019.

\bibitem{bromley1994signature}
J.~Bromley, I.~Guyon, Y.~LeCun, E.~S{\"a}ckinger, and R.~Shah, ``Signature
  verification using a ``siamese" time delay neural network,'' in
  \emph{Advances in neural information processing systems}, 1994, pp. 737--744.

\bibitem{huber2004robust}
P.~J. Huber, \emph{Robust statistics}.\hskip 1em plus 0.5em minus 0.4em\relax
  John Wiley \& Sons, 2004, vol. 523.

\bibitem{bolles1981ransac}
R.~C. Bolles and M.~A. Fischler, ``A ransac-based approach to model fitting and
  its application to finding cylinders in range data.'' in \emph{IJCAI}, vol.
  1981, 1981, pp. 637--643.

\bibitem{dang2008theil}
X.~Dang, H.~Peng, X.~Wang, and H.~Zhang, ``Theil-sen estimators in a multiple
  linear regression model,'' \emph{Olemiss Edu}, 2008.

\bibitem{scikit-learn}
F.~Pedregosa, G.~Varoquaux, A.~Gramfort, V.~Michel, B.~Thirion, O.~Grisel,
  M.~Blondel, P.~Prettenhofer, R.~Weiss, V.~Dubourg, J.~Vanderplas, A.~Passos,
  D.~Cournapeau, M.~Brucher, M.~Perrot, and E.~Duchesnay, ``Scikit-learn:
  Machine learning in {P}ython,'' \emph{Journal of Machine Learning Research},
  vol.~12, pp. 2825--2830, 2011.

\bibitem{nelder1965simplex}
J.~A. Nelder and R.~Mead, ``A simplex method for function minimization,''
  \emph{The computer journal}, vol.~7, no.~4, pp. 308--313, 1965.

\bibitem{wang2020directshape}
R.~Wang, N.~Yang, J.~St{\"u}ckler, and D.~Cremers, ``Directshape: Direct
  photometric alignment of shape priors for visual vehicle pose and shape
  estimation,'' in \emph{2020 IEEE International Conference on Robotics and
  Automation (ICRA)}.\hskip 1em plus 0.5em minus 0.4em\relax IEEE, 2020, pp.
  11\,067--11\,073.

\bibitem{xie2017aggregated}
S.~Xie, R.~Girshick, P.~Doll{\'a}r, Z.~Tu, and K.~He, ``Aggregated residual
  transformations for deep neural networks,'' in \emph{Proceedings of the IEEE
  conference on computer vision and pattern recognition}, 2017, pp. 1492--1500.

\bibitem{he2017mask}
K.~He, G.~Gkioxari, P.~Doll{\'a}r, and R.~Girshick, ``Mask r-cnn,'' in
  \emph{Proceedings of the IEEE international conference on computer vision},
  2017, pp. 2961--2969.

\bibitem{weng20203d}
X.~Weng, J.~Wang, D.~Held, and K.~Kitani, ``3d multi-object tracking: A
  baseline and new evaluation metrics,'' \emph{arXiv preprint
  arXiv:1907.03961}, 2020.

\bibitem{brazil2020kinematic}
G.~Brazil, G.~Pons-Moll, X.~Liu, and B.~Schiele, ``Kinematic 3d object
  detection in monocular video,'' \emph{arXiv preprint arXiv:2007.09548}, 2020.

\end{thebibliography}

\end{document}